\documentclass[10pt,twocolumn,letterpaper]{article}

\usepackage{iccv}
\usepackage{times}
\usepackage{epsfig}
\usepackage{graphicx}
\usepackage{amsmath, bm}
\usepackage{amssymb}
\usepackage[misc]{ifsym} 
\usepackage{caption}
\usepackage{subcaption}
\newcommand*{\affmark}[1][*]{\textsuperscript{#1}}

\usepackage[breaklinks=true,bookmarks=false]{hyperref}

\iccvfinalcopy 

\def\iccvPaperID{1583} 

\hypersetup{
	colorlinks=true,
	linkcolor=red,
}

\ificcvfinal\fi
\begin{document}
	
\title{Expectation-Maximization Attention Networks for Semantic Segmentation}
\author{Xia Li\affmark[1,2],\quad Zhisheng Zhong\affmark[2],\quad Jianlong Wu\affmark[2,3],\quad Yibo Yang\affmark[2,4],\quad Zhouchen Lin\affmark[2],\quad Hong Liu\affmark[1,]\affmark[\Letter]\\
\small \affmark[1] Key Laboratory of Machine Perception, Shenzhen Graduate School, Peking University\\
\small \affmark[2] Key Laboratory of Machine Perception (MOE), School of EECS, Peking University\\
\small \affmark[3] School of Computer Science and Technology, Shandong University\\
\small \affmark[4] Academy for Advanced Interdisciplinary Studies, Peking University\\
{\tt\small \{{ethanlee,zszhong,jlwu1992,ibo,zlin,hongliu\}@pku.edu.cn}}}
	
\maketitle
\ificcvfinal\thispagestyle{empty}\fi

\begin{abstract}
	Self-attention mechanism has been widely used for various tasks. It is designed to compute the representation of each position by a weighted sum of the features at all positions. Thus, it can capture long-range relations for computer vision tasks. However, it is computationally consuming. Since the attention maps are computed w.r.t all other positions. In this paper, we formulate the attention mechanism into an expectation-maximization manner and iteratively estimate a much more compact set of bases upon which the attention maps are computed. By a weighted summation upon these bases, the resulting representation is low-rank and deprecates noisy information from the input. The proposed \textbf{E}xpectation-\textbf{M}aximization \textbf{A}ttention~(EMA) module is robust to the variance of input and is also friendly in memory and computation. Moreover, we set up the bases maintenance and normalization methods to stabilize its training procedure. We conduct extensive experiments on popular semantic segmentation benchmarks including PASCAL VOC, PASCAL Context and COCO Stuff, on which we set new records\footnote{Project address: \url{https://xialipku.github.io/EMANet}}.
\end{abstract}
	
\section{Introduction}
	
Semantic segmentation is a fundamental and challenging problem of computer vision, whose goal is to assign a semantic category to each pixel of the image. It is critical for various tasks such as autonomous driving, image editing and robot sensing. In order to accomplish the semantic segmentation task effectively, we need to distinguish some confusing categories and take the appearance of different objects into account. For example, `grass' and `ground' have similar color in some cases and `person' may have various scales, figures and clothes in different locations of the image. Meanwhile, the label space of the output is quite compact and the amount of the categories for a specific dataset is limited. Therefore, this task can be treated as projecting data points in a high-dimensional noisy space into a compact sub-space. The essence lies in de-noising these variation and capturing the most important semantic concepts.
	
Recently, many state-of-the-art methods based on fully convolutional networks~(FCNs)~\cite{fcn} have been proposed to address the above issues. Due to the fixed geometric structures, they are inherently limited by local receptive  fields and short-range contextual information. To capture long-range dependencies, several works employ the multi-scale context fusion~\cite{rescan}, such as astrous convolution~\cite{deeplabv3}, spatial pyramid~\cite{pspnet}, large kernel convolution~\cite{gcn} and so on. Moreover, to keep more detailed information, the encoder-decoder structures~\cite{dfn,deeplabv3+} are proposed to fuse mid-level and high-level semantic features. To aggregate information from all spatial locations, attention mechanism~\cite{attention,psanet,nonlocal} is used, which enables the feature of a single  pixel to fuse information from all other positions. However, the original attention-based methods need to generate a large attention map, which has high computation complexity and occupies a huge number of GPU memory. The bottleneck lies in that both the generation of attention map and its usage are computed w.r.t all positions.
	
\begin{figure}
	\centering
	\includegraphics[width=0.48\textwidth]{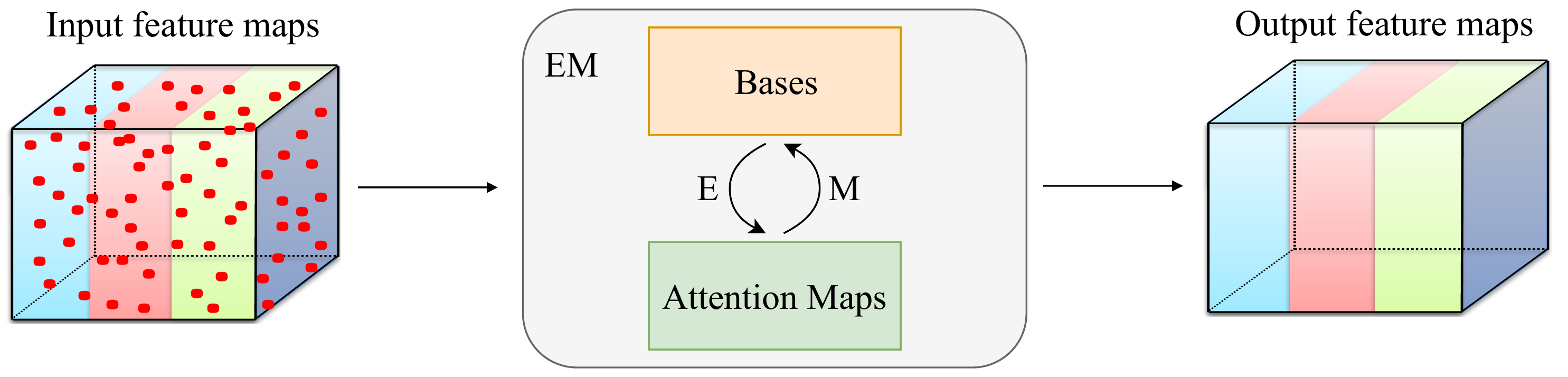}
	\vspace{-5mm}
	\caption{Pipeline of the proposed expectation-maximization attention method.}
	\label{fig-pipeline}
	\vspace{-5mm}
\end{figure}

	
Towards the above issues, in this paper, we rethink the attention mechanism from the view of expectation-maximization~(EM) algorithm~\cite{em} and propose a novel attention-based method, namely \textbf{E}xpectation-\textbf{M}aximization \textbf{A}ttention~(EMA). 
Instead of treating all pixels themselves as the reconstruction bases~\cite{psanet,nonlocal}, we use the EM algorithm to find a more compact basis set, which can largely reduce the computational complexity. In detail, we regard the bases for construction as the parameters to learn in the EM algorithm and attention maps as latent variables. In this setting, the EM algorithm aims to find a maximum likelihood estimate of parameters~(bases). Given the current parameters, the expectation~(E) step works as estimating the expectation of attention map and maximization~(M) step functions as updating the parameters~(bases) by maximizing the complete data likelihood. The E step and the M step execute alternately. After convergence, the output can be computed as the weighted sum of bases, where the weights are the normalized final attention maps. The pipeline of EMA is shown in Fig.~\ref{fig-pipeline}.
	
We further embed the proposed EMA method into a module for neural network, which is named EMA Unit. EMA Unit can be simply implemented by common operators. It is also light-weighted and can be easily embedded into existing neural networks. Moreover, to make full use of its capacity, we also propose two more methods to stabilize the training process of EMA Unit. We also evaluate its performance on three challenging datasets.
	
	The main contributions of this paper are listed as follows:
	
	\begin{itemize}
	\item We reformulate the self-attention mechanism into an expectation-maximization iteration manner, which can learn a more compact basis set and largely reduce the computational complexity. To the best of our knowledge, this paper is the first to introduce EM iterations into attention mechanism.
	
	\item We build the proposed expectation-maximization attention as a light-weighted module for neural network and set up specific manners for bases' maintenance and normalization.
	
	\item Extensive experiments on three challenging semantic segmentation datasets, including PASCAL VOC, PASCAL Context and COCO Stuff, demonstrate the superiority of our approach over other state-of-the-art methods.
	\end{itemize}
	
	\section{Related Works}
	
	\noindent\textbf{Semantic segmentation.} Fully convolutional network (FCN)~\cite{fcn} based methods have made great progress in image semantic segmentation by leveraging the powerful convolutional features of classification networks~\cite{resnet,densenet,cliquenet} pre-trained on large-scale data~\cite{imagenet}. Several model variants are proposed to enhance the multi-scale contextual aggregation. For example, DeeplabV2~\cite{deeplabv3} makes use of the astrous spatial pyramid pooling~(ASPP) to embed contextual information, which consists of parallel dilated convolutions with different dilated rates. DeeplabV3~\cite{deeplabv3} extends ASPP with image-level feature to further capture global contexts. Meanwhile, PSPNet~\cite{pspnet} proposes a pyramid pooling module to collect contextual information of different scales. GCN~\cite{gcn} adopts decoupling of large kernel convolution to gain a large receptive field for the feature map and capture long-range information.
	
	For the other type of variants, they mainly focus on predicting more detailed output. These methods are based on U-Net~\cite{unet}, which combines the advantages of high-level features with mid-level features. RefineNet~\cite{refinenet} makes use of the Laplacian image pyramid to explicitly capture the information available along the down-sampling process and output predictions from coarse to fine. DeeplabV3+~\cite{deeplabv3+} adds a decoder upon DeeplabV3 to refine the segmentation results especially along object boundaries. Exfuse~\cite{exfuse} proposes a new framework to bridge the gap between low-level and high-level features and thus improves the segmentation quality.
	
	\noindent\textbf{Attention model.}
	Attention is widely used for various tasks such as machine translation, visual question answering and video classification. The self-attention methods~\cite{nmt,attention} calculate the context coding at one position by a weighted summation of embeddings at all positions in sentences. Non-local~\cite{nonlocal} first adopts self-attention mechanism as a module for computer vision tasks, such as video classification, object detection and instance segmentation. PSANet~\cite{psanet} learns to aggregate contextual information for each position via a predicted attention map. $A^2$Net~\cite{a2net} proposes the double attention block to distribute and gather informative global features from the entire spatio-temporal space of the images. DANet~\cite{danet} applies both spatial and channel attention to gather information around the feature maps, which costs even more computation and memory than the Non-local method.
	
	Our approach is motivated by the success of attention in the above works. We rethink the attention mechanism from the view of the EM algorithm and compute the attention map in an iterative manner as the EM algorithm. 
	
	\section{Preliminaries}
	
	Before introducing our proposed method, we first review three highly correlated methods, that is the EM algorithm, the Gaussian mixture model and the Non-local module.
	
	\subsection{Expectation-Maximization Algorithm}
	
	The expectation-maximization~(EM)~\cite{em} algorithm aims to find the maximum likelihood solution for latent variables models. Denote $\mathbf{X} = \left\{ \mathbf{x}_1, \mathbf{x}_2, \cdots, \mathbf{x}_N \right\}$ as the data set which consists of $N$ observed samples and each data point $\mathbf{x}_i$ has its corresponding latent variable $\mathbf z_i$. We  call $\left\{ \mathbf{X}, \mathbf{Z} \right\}$ the complete data and its likelihood function takes the form $\ln p \left( \mathbf{X}, \mathbf{Z} \vert \bm{\theta} \right)$, where $\bm{\theta}$ is the set of all parameters of the model. In practice, the only knowledge of latent variables in $\mathbf{Z}$ is given by the posterior distribution $p \left( \mathbf{Z} \vert \mathbf{X}, \bm{\theta} \right)$. The EM algorithm is designed to maximize the likelihood $\ln p \left( \mathbf{X}, \mathbf{Z} \vert \bm{\theta} \right)$ by two steps, i.e., the E step and the M step. 
	
	In the E step, we use the current parameters $\bm{\theta}^{\mathrm{old}}$ to find the posterior distribution of $\mathbf{Z}$ given by $p \left( \mathbf{X}, \mathbf{Z} \vert \bm{\theta} \right)$. Then we use the posterior distribution to find the expectation of the complete data likelihood $\mathcal{Q} \left( \bm{\theta}, \bm{\theta}^{\mathrm{old}} \right)$, which is given by:
	\begin{equation}
	\mathcal{Q} \left( \bm{\theta}, \bm{\theta}^{\mathrm{old}} \right) = \sum_{\mathbf z} p \left( \mathbf{Z} \vert \mathbf{X}, \bm{\theta}^{\mathrm{old}} \right) \ln p \left( \mathbf{X}, \mathbf{Z} \vert \bm{\theta} \right).
	\end{equation}
	Then in the M step, the revised parameter $\bm{\theta}^{\mathrm{new}}$ is determined by maximizing the function:
	\begin{equation}
	\bm{\theta}^{\mathrm{new}} = \underset{\bm{\theta}}{\arg\max} \mathcal{Q} \left( \bm{\theta}, \bm{\theta}^{\mathrm{old}} \right).
	\end{equation}
	The EM algorithm executes the E step and the M step alternately until the convergence criterion is satisfied.
	
	\subsection{Gaussian Mixture Model}
	
	Gaussian mixture model~(GMM)~\cite{gmm} is a special case of the EM algorithm. It takes the distribution of data $\mathbf{x}_n$ as a linear superposition of Gaussians:
	\begin{equation}
	p\left( \mathbf{x}_n \right) = \sum_{k=1}^{K} z_{nk} \mathcal{N} \left( \mathbf{x}_n \vert \bm{\mu}_k, \mathbf{\Sigma}_k \right),
	\end{equation}
	where the mean $\bm{\mu}_k$ and the covariance $\mathbf{\Sigma}_k$ are parameters for the $k$-th Gaussian basis.
	Here we leave out the prior $\pi_k$.
	The likelihood of the complete data is formulated as:
	\begin{equation}
	\ln p \left( \mathbf{X}, \mathbf{Z} \vert \bm{\mu}, \bm{\Sigma} \right) = \sum_{n=1}^{N} \ln \left[ \sum_{k=1}^{K} z_{nk} \mathcal{N} \left( \mathbf{x}_n \vert \bm{\mu}_k, \mathbf{\Sigma}_k \right) \right],
	\label{eq-lnp}
	\end{equation}
	where $\sum_{k} z_{nk} = 1$. $z_{nk}$ can be viewed as the responsibility that the $k$-th basis takes for the observation $\mathbf{x}_n$. For GMM, in the E step, the expected value of $z_{nk}$ is given by:
	\begin{equation}
	z_{nk}^{\mathrm{new}} = \frac{ \mathcal{N} \left( \mathbf{x}_n \vert \bm{\mu}_k^{\mathrm{new}}, \mathbf{\Sigma}_k \right) }{ \sum_{j=1}^{K} \mathcal{N} \left( \mathbf{x}_n \vert \bm{\mu}_j^{\mathrm{new}}, \mathbf{\Sigma}_j \right )}.
	\label{eq-znk}
	\end{equation}
	In the M step, the parameters are re-estimated as follows:
	\begin{equation}
	\begin{aligned}
	\bm{\mu}_{k}^{\mathrm{new}} &= \frac{1}{N_k} \sum_{n=1}^{N} z_{nk}^{\mathrm{new}} \mathbf{x}_n, \\\
	\bm{\Sigma}_{k}^{\mathrm{new}} &= \frac{1}{N_k} \sum_{n=1}^{N} z_{nk}^{\mathrm{new}} \left( \mathbf{x}_n - \bm{\mu}_{k}^{\mathrm{old}} \right) \left( \mathbf{x}_n - \bm{\mu}_{k}^{\mathrm{old}} \right)^{\top},
	\end{aligned}
	\end{equation} 
	where
	\begin{equation}
	N_k = \sum_{n=1}^{N} z_{nk}^{\mathrm{new}}.
	\end{equation}
	After the convergence of the GMM parameters, the re-estimated $\mathbf{x}_{n}^{\mathrm{new}}$ can be formulated as:
	\begin{equation}
	\mathbf{x}_{n}^{\mathrm{new}} = \sum_{k=1}^{K} z_{nk}^{\mathrm{new}} \bm{\mu}_k^{\mathrm{new}}.
	\label{eq-recon}
	\end{equation}
	
	In real applications, we can simply replace $\mathbf{\Sigma}_{k}$ as the identity matrix $\mathbf{I}$ and leave out the $\mathbf{\Sigma}_k$ in the above equations.
	
	\subsection{Non-local}
	
	The Non-local module~\cite{nonlocal} functions the same as the self-attention mechanism. It can be formulated as:
	\begin{equation}
	\mathbf{y}_i = \frac{1}{ \mathcal{C} \left( \mathbf{x}_i  \right) } \sum_j f \left( \mathbf{x}_i, \mathbf{x}_j \right) g \left( \mathbf{x}_j \right),
	\label{eq-nonlocal}
	\end{equation}
	where $f \left(\cdot, \cdot\right)$ represents a general kernel function, $\mathcal{C} \left( \mathbf{x}  \right)$ is a normalization factor and $\mathbf{x}_i$ denotes the feature vector for the location $i$.
	As this module is applied upon the feature map of convolutional neural networks~(CNN).
	
	Considering that $\mathcal{N} \left( \mathbf{x}_n \vert \bm{\mu}_k, \mathbf{\Sigma}_k \right) $ in Eq.~(\ref{eq-znk}) is a specific kernel function between $\mathbf{x}_n$ and $\bm{\mu}_k$, Eq.~(\ref{eq-recon}) is just a specific design of Eq.~(\ref{eq-nonlocal}). Then, from the viewpoint of GMM, the Non-local module is just a re-estimation of $\mathbf{X}$, without E steps and M steps. Specifically,  $\bm{\mu}$ is just selected as the $\mathbf{X}$ in Non-local.
	
	In GMM, the number of Gaussian bases is selected manually and usually satisfies $K \ll N$. But in the Non-local module, the bases are selected as the data themselves, so it has $K = N$. 
	There are two obvious disadvantages of the Non-local module. 
	First, the data are lying in a low dimensional manifold, so the bases are over-complete. Second, the computation overhead is heavy and the memory cost is also large. 
	
	\section{Expectation-Maximization Attention}
	
	\begin{figure*}[ht!]
		\centering
		\includegraphics[width=1\textwidth]{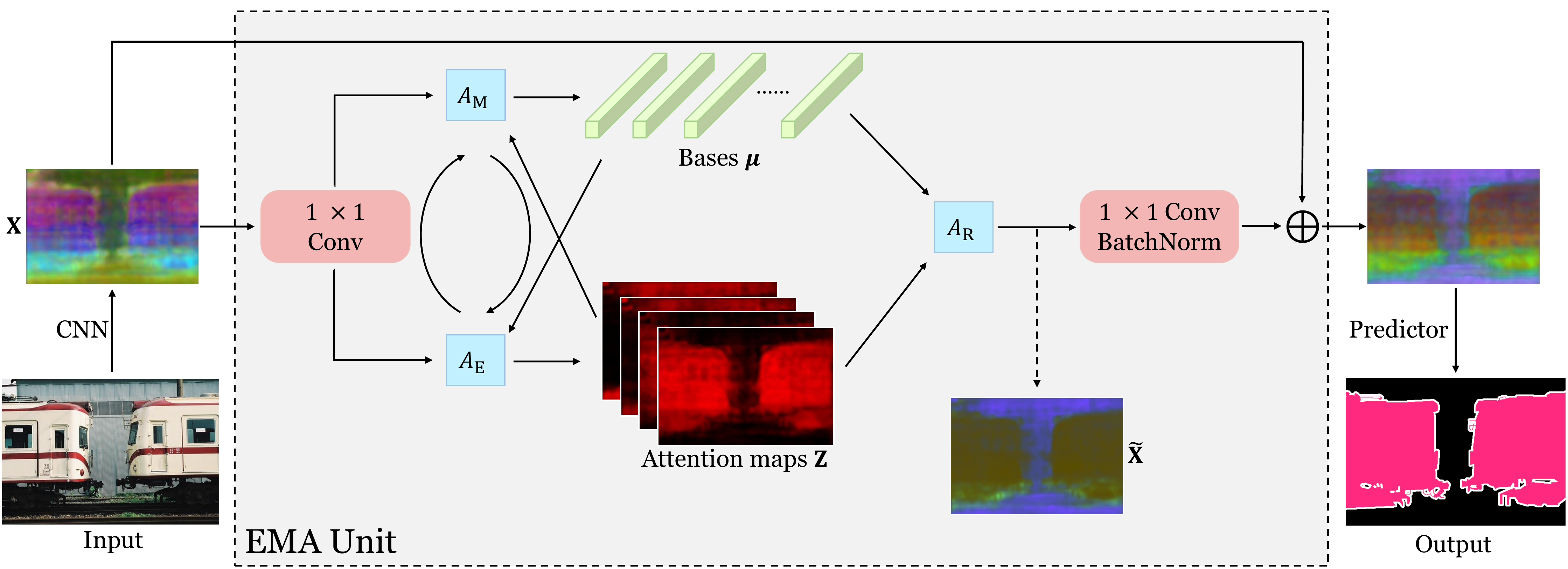}
		\vspace{-5mm}
		\caption{Overall structure of the proposed EMAU. The key component is the EMA operator, in which $A_{\mathrm{E}}$ and $A_{\mathrm{M}}$ execute alternately. In addition to the EMA operator, we add two $1 \times 1$ convolutions at the beginning and the end of EMA and sum the output with original input, to form a residual-like block. \textit{Best viewed on screen}.}
		\label{fig-structure}
		\vspace{-5mm}
	\end{figure*}
	In view of the high computational complexity of the attention mechanism and limitations of the Non-local module, we first propose the expectation-maximization attention~(EMA) method, which is an augmented version of self-attention. Unlike the Non-local module that selects all data points as bases, we use the EM iterations to find a compact basis set.
	
	For simplicity, we consider an input feature map $\mathbf{X}$ of size $C \times H \times W$ from a single sample. $\mathbf{X}$ is the intermediate activations of a CNN. To simplify the symbols, we reshape $\mathbf{X}$ into $N \times C$, where $N = H \times W$, and $\mathbf{x}_i \in \mathbb{R}^C$ indexes the $C$ dimensional feature vector at pixel $i$. Our proposed EMA consists of three operations, including \textbf{responsibility estimation}~($A_{\mathrm E}$), \textbf{likelihood maximization}~($A_{\mathrm M}$) and \textbf{data re-estimation}~($A_{\mathrm R}$). Briefly, given the input $\mathbf{X} \in \mathbb{R}^{N \times C}$ and the initial bases $\bm{\mu} \in \mathbb{R}^{K \times C}$, $A_{\mathrm{E}}$ estimates the latent variables~(or the `responsibility') $\mathbf{Z} \in \mathbb{R}^{N \times K}$, so it functions as the E step in the EM algorithm. $A_{\mathrm{M}}$ uses the estimation to update the bases $\bm{\mu}$, which works as the M step. The $A_{\mathrm{E}}$ and $A_{\mathrm{M}}$ steps execute alternately for a pre-specified number of iterations. Then, with the converged $\bm{\mu}$ and $\mathbf{Z}$, $A_{\mathrm{R}}$ reconstructs the original $\mathbf{X}$ as $\mathbf{Y}$ and outputs it.
	
	It has been proved that, with the iteration of EM steps, the complete data likelihood $\ln p \left( \mathbf{X}, \mathbf{Z} \right)$ will increase monotonically. As $\ln p \left( \mathbf{X} \right)$ can be estimated by marginalizing $\ln p \left( \mathbf{X}, \mathbf{Z} \right)$ with $\mathbf{Z}$, maximizing $\ln p \left( \mathbf{X}, \mathbf{Z} \right)$ is a proxy of maximizing $\ln p \left( \mathbf{X} \right)$. Therefore, with the iterations of  $A_{\mathrm{E}}$ and $A_{\mathrm{M}}$, the updated $\mathbf{Z}$ and $\bm{\mu}$ have better ability to reconstruct the original data $\mathbf{X}$. The reconstructed $\tilde{\mathbf{X}}$ can capture important semantics from $\mathbf{X}$ as much as possible.
	
	Moreover, compared with the Non-local module, EMA finds a compact set of bases for pixels of an input image. The compactness is non-trivial. Since $K \ll N$, $\tilde{\mathbf{X}}$ lies in a subspace of $\mathbf{X}$. This mechanism removes much unnecessary noise and makes the final classification of each pixel more tractable. Moreover, this operation reduces the complexity~(both in space and time) from $O \left(N^2\right)$ to $O \left(NKT\right)$, where $T$ is the  number of iterations for $A_{\mathrm{E}}$ and $A_{\mathrm{M}}$. The convergence of EM algorithm is also guaranteed. Notably, EMA takes only three iterations to get promising results in our experiments. So $T$ can be treated as a small constant, which means that the complexity is only $O \left(NK\right)$.
	
	\subsection{Responsibility Estimation}
	
	Responsibility estimation~($A_{\mathrm{E}}$) functions as the E step in the EM algorithm. This step computes the expected value of $z_{nk}$, which corresponds to the responsibility of the $k$-th basis $\bm{\mu}$ to $\mathbf{x}_n$, where $1 \leq k \leq K$ and $1 \leq n \leq N$. We formulate the posterior probability of $\mathbf{x}_n$ given $\bm{\mu}_k$ as follows:
	\begin{equation}
	p \left( \mathbf{x}_n \vert \bm{\mu}_k \right) = \mathcal{K} \left( \mathbf{x}_n, \bm{\mu}_k \right),
	\end{equation}
	where $\mathcal{K}$ represents the general kernel function. And now, Eq.~(\ref{eq-znk}) can be reformulated into a more general form:
	\begin{equation}
	z_{nk} = \frac{ \mathcal{K} \left( \mathbf{x}_n, \bm{\mu}_k \right) }{ \sum_{j=1}^{K} \mathcal{K} \left( \mathbf{x}_n, \bm{\mu}_j \right) }.
	\label{eq-AE}
	\end{equation}
	There are several choices for $\mathcal{K} \left( \mathbf{a}, \mathbf{b} \right)$, such as inner dot $\mathbf{a}^{\top}\mathbf{b}$, exponential inner dot $\exp \left( \mathbf{a}^{\top}\mathbf{b}\right)$, Euclidean distance $\left\| \mathbf{a} -\mathbf{b}\right\|_2^2$, RBF kernel $\exp \left( - \left\|  \mathbf{a} -\mathbf{b}\right\|_2^2 / \sigma^2 \right)$ and so on. As compared in the Non-local module, the choice of these functions makes trivial differences in the final results. So we simply take the exponential inner dot $\exp\left(\mathbf{a}^{\top}\mathbf{b}\right)$ in our paper. In experiments, Eq.~(\ref{eq-AE}) can be implemented as a matrix multiplication plus one softmax layer. In conclusion, the operation of $A_{\mathrm{E}}$ in the $t$-th iteration is formulated as:
	\begin{equation}
	\mathbf{Z}^{(t)} = \mathrm{softmax} \left( \lambda \mathbf{X} \left( \bm{\mu}^{(t-1)} \right)^{\top} \right),
	\label{eq-ae}
	\end{equation}
	where $\lambda$ is a hyper-parameter to control the distribution of $\mathbf{Z}$.
	
	\subsection{Likelihood Maximization}
	
	Likelihood maximization~($A_{\mathrm{M}}$) works as the EM algorithm's M step. With the estimated $\mathbf{Z}$, $A_{\mathrm{M}}$ updates $\bm{\mu}$ by maximizing the complete data likelihood. To keep the bases lying in the same embedding space as $\mathbf{X}$, we update the bases $\bm{\mu}$ using the weighted summation of $\mathbf{X}$. So $\bm{\mu}_k$ is updated as
	\begin{equation}
	\bm{\mu}_k^{(t)} = \frac{z_{nk}^{(t)} \mathbf{x}_n }{ \sum_{m=1}^{N} z_{mk}^{(t)} }
	\end{equation}
	in the $t$-th iteration of $A_{\mathrm{M}}$.
	
	It is noteworthy that if we set $\lambda \to \infty$ in Eq.~(\ref{eq-ae}), then $\left\{ z_{n1}, z_{n2}, \cdots, z_{nK} \right\}$ will become a one-hot embedding. In this situation, each pixel is assigned to only one basis. And the basis is updated by the average of those pixels assigned to it. This is what the K-means clustering algorithm~\cite{kmeans} does. So the iterations of $A_{\mathrm{E}}$ and $A_{\mathrm{M}}$ can also be viewed as a soft version of K-means clustering.
	
	\subsection{Data Re-estimation}
	
	EMA runs $A_{\mathrm{E}}$ and $A_{\mathrm{M}}$ alternately for $T$ times. After that, the final $\bm{\mu}^{\left(T\right)}$ and $\mathbf{Z}^{\left(T\right)}$ are used to re-estimate the $\mathbf{X}$. We adopt Eq.~(\ref{eq-recon}) to construct the new $\mathbf{X}$, namely $\tilde{\mathbf{X}}$, which is formulated as:
	\begin{equation}
	\tilde{\mathbf{X}} = \mathbf{Z}^{(T)} \bm{\mu}^{(T)}.
	\end{equation}
	As $\tilde{\mathbf{X}}$ is constructed from a compact basis set, it has the low-rank property compared with the input $\mathbf{X}$. We depict an example of $\tilde{\mathbf{X}}$ in Fig.~\ref{fig-structure}. It's obvious that  $\tilde{\mathbf{X}}$ outputed from $A_{\mathrm{R}}$ is very compact in the feature space and the feature variance inside the object is smaller than that of the input. 
	
	\section{EMA Unit}
	In order to better incorporate the proposed EMA with deep neural networks,
	we further propose the Expectation-maximization Attention Unit~(EMAU) and apply it to semantic segmentation task. In this section, we will describe EMAU in detail. We first introduce the overall structure of EMAU and then discuss bases' maintenance and normalization mechanisms.
	
	\subsection{Structure of EMA Unit}
	
	The overall structure of EMAU is shown in Fig.~\ref{fig-structure}. EMAU looks like the bottleneck of ResNet at the first glance, except it replaces the heavy $3 \times 3$ convolution with the EMA operations. The first convolution without the ReLU activation is prepended to transform the value range of input from $\left(0, +\infty\right)$ to $\left( -\infty, +\infty \right)$. This transformation is very important, or the estimated $\bm{\mu}^{\left( T \right)}$ will also lie in $[ 0, +\infty )$, which halves the capacity compared with general convolution parameters. The last $1 \times 1$ convolution is inserted to transform the re-estimated $\tilde{\mathbf{X}}$ into the residual space of $\mathbf{X}$.
	
	For each of $A_{\mathrm{E}}$, $A_{\mathrm{M}}$ and $A_{\mathrm{R}}$ steps, the computation complexity is $O \left( NKC \right)$. As we set $K \ll C$, several iterations of $A_{\mathrm{E}}$ and $A_{\mathrm{M}}$ plus one $A_{\mathrm{R}}$ is just the same magnitude as a $1 \times 1$ convolution with input and output channel numbers all being $C$. Adding the extra computation from two $1 \times 1$ convolutions, the whole FLOPs of EMAU is around $1/3$ of a module running $3 \times 3$ convolutions with the same number of input and output channels. Moreover, the parameters maintained by EMA just counts to $KC$.
	
	\subsection{Bases Maintenance}\label{sec-comp_init}
	
	Another issue for the EM algorithm is the initialization of the bases. The EM algorithm is guaranteed to converge, because the likelihood of complete data is limited, and at each iteration both E and M steps lift its current lower bound. However, converging to global maximum is not guaranteed. Thus, the initial values of bases before iterations are of great importance.
	
	We only describe how EMA is used to process one image above. However, for a computer vision task, there are thousand of images in a dataset. As each image $\mathbf{X}$ has different pixel feature distributions from others, it is not suitable to use the $\bm{\mu}$ computed upon an image to reconstruct feature maps of other images. So we run EMA on each image. 
	
	For the first mini-batch, we initialize $\bm{\mu}^{(0)}$ using Kaiming's initialization~\cite{kaiming}, where we treat matrix multiplication as a $1 \times 1$ convolution.
	For the following mini-batches, one simple choice is to update $\bm{\mu}^{(0)}$ using standard back propagation. However, as iterations of $A_E$ and $A_M$ can be unrolled as a recurrent neural network~(RNN), the gradients propagating though them will encounter the vanishing or explosion problem. Therefore, the updating of $\bm{\mu}^{(0)}$ is unstable, and the training procedure of EMA Unit may collapse.
	
	In this paper, we use the moving averaging to update $\bm{\mu}^{(0)}$ in the training process. After iterating over an image, the generated $\bm{\mu}^{(T)}$ can be regarded as a biased update of $\bm{\mu}^{(0)}$, where the bias comes from the image sampling process. To make it less biased, we first average $\bm{\mu}^{(T)}$ over a mini-batch and get the $\bm{\bar{\mu}}^{(T)}$. Then we update $\bm{\mu}^{(0)}$ as:
	\begin{equation}
	\bm{\mu}^{(0)} \leftarrow \alpha \bm{\mu}^{(0)} + \left( 1 - \alpha \right) \bm{\bar{\mu}}^{(T)},
	\end{equation}
	where $\alpha \in \left[ 0, 1 \right]$ is the momentum. For inference, the $\bm{\mu}^{(0)}$ keeps fixed. This moving averaging mechanism is also adopted in batch normalization (BN) ~\cite{bn}.

	\subsection{Bases Normalization}\label{sec-comp_norm}

	In the above subsection, we accomplish the maintenance of $\bm{\mu}^{(0)}$ for each mini-batch. However, the stable update of $\bm{\mu}^{(t)}$ inside $A_{\mathrm{E}}$ and $A_{\mathrm{M}}$ iterations is still not guaranteed, due to the defect of RNN. The moving averaging mechanism described above requires $\bm{\bar{\mu}}^{(T)}$ not to differ significantly from $\bm{\mu}^{(0)}$, otherwise it will also collapse like back-propagation. This requirement also constrains the value range of $\bm{\mu}^{(t)}, 1 \leq t \leq T$. 
	
	To this end, we need to apply normalization upon $\bm{\mu}^{(t)}$. At the first glance, BN or layer normalization~(LN)~\cite{ln} sound to be good choices.
	However, these aforementioned normalization methods will change the direction of each basis $\bm{\mu}_k^{(t)}$, which changes their properties and semantic meanings. To keep the direction of each basis untouched, we choose Euclidean normalization~(L2Norm), which divides each $\bm{\mu}_k^{(t)}$ by its length. By applying it, $\bm{\mu}^{(t)}$ then lies in a $K$-dimensional united hyper-sphere, and sequence of $\left\{ \bm{\mu}_k^{(0)}, \bm{\mu}_k^{(1)}, \cdots, \bm{\mu}_k^{(T)} \right\}$ forms a trajectory on it.

	\subsection{Compare with the Double Attention Block}\label{sec-a2net}
	
	$A^2$ Net~\cite{a2net} proposes the double attention block ($A^2$ block), in which the output $\mathbf{Y}$ is computed as:
	\begin{equation}
	\mathbf{Y}\!=\left[\phi\!\left(\!\mathbf{X}, W_{\phi} \right) \mathrm{sfm}\!\left(\theta \left( \mathbf{X}, W_{\theta}\right)\right)^\top\!\right]\!\mathrm{sfm}\!\left( \rho \left( \mathbf{X}, W_{\rho} \right) \right)\!,
	\end{equation}
	where $\mathrm{sfm}$ represents the $\mathrm{softmax}$ function. $\phi$, $\theta$ and $\rho$ represent three $1 \times 1$ convolutions with convolution kernels $W_{\phi}$, $W_{\theta}$ and $W_{\rho}$, respectively.
	
	If we share parameters between $\theta$ and $\rho$, then we can mark both $W_{\theta}$ and $W_{\rho}$ as $\bm{\mu}$. We can see  that $\mathrm{sfm} \left( \theta \left( \mathbf{X}, W_{\theta} \right) \right)$ just computes $\mathbf{Z}$ the same as Eq.~(\ref{eq-znk}) and those variables lying inside $\left[ \cdot \right]$ update $\bm{\mu}$. The whole process of $A^2$ block equals to EMA with only one iteration. The $W_{\theta}$ in $A^2$ block is updated by the back-propagation, while our EMAU is updated by moving averaging. Above all, double attention block can be treated as a special form of EMAU.
	
	\begin{figure}
		\centering
		\begin{subfigure}{0.23\textwidth}
			\includegraphics[width=\textwidth]{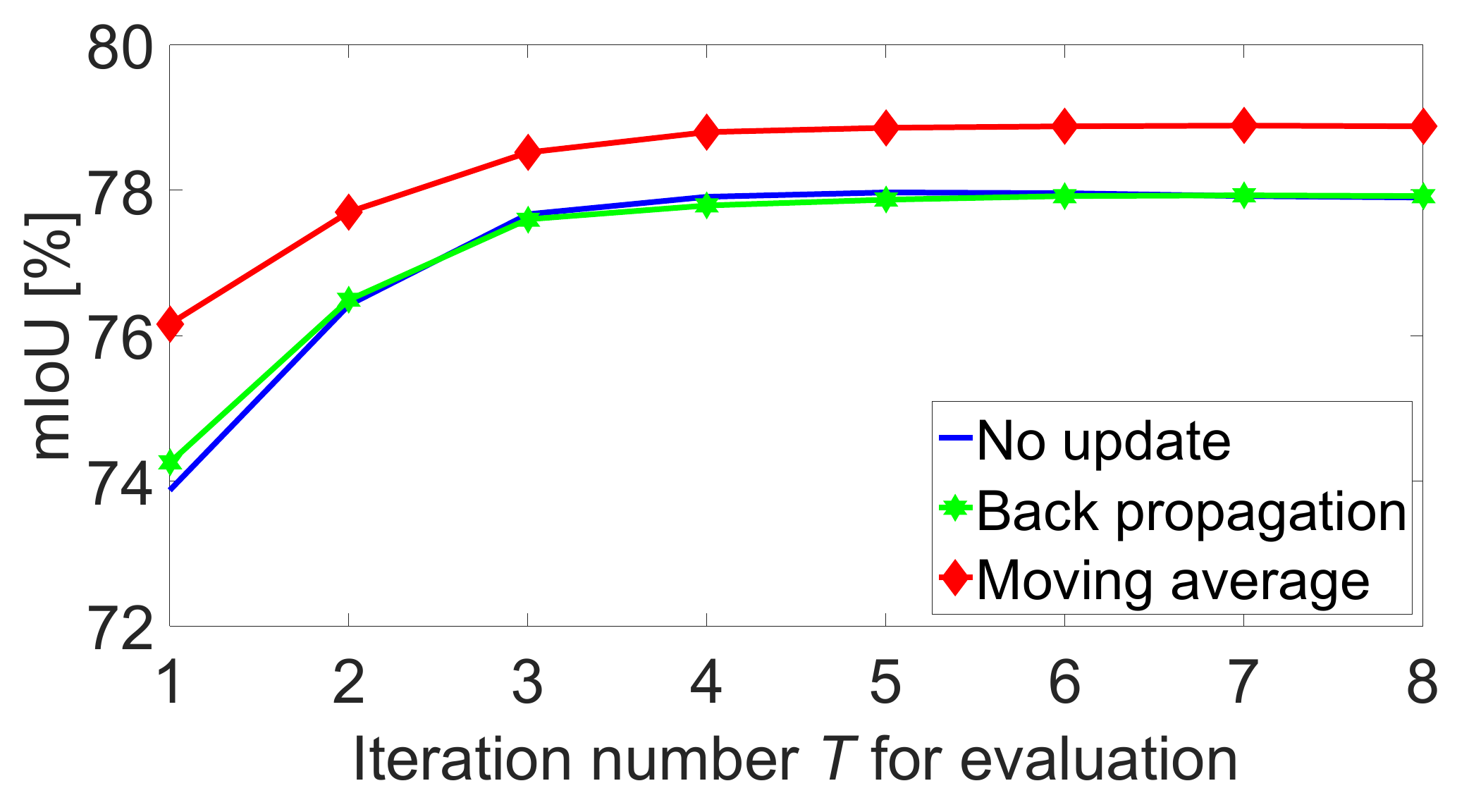}
		\end{subfigure}
		\begin{subfigure}{0.23\textwidth}
			\includegraphics[width=\textwidth]{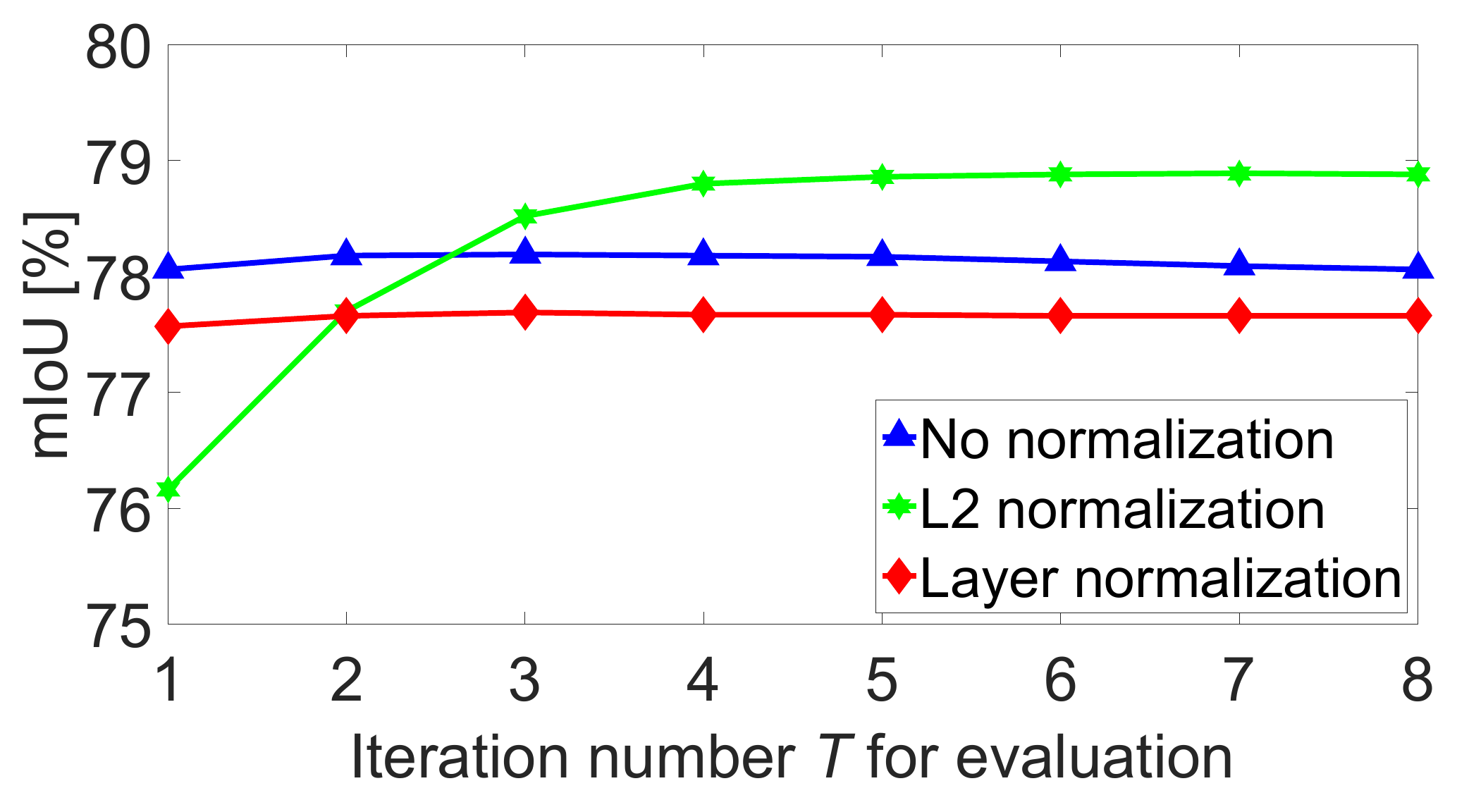}
		\end{subfigure}
		\vspace{-2mm}
		\caption{Ablation study on strategy of bases maintenance (left) and normalization (right) of EMAU. Experiments are carried out upon ResNet-50 with batch size $12$ and training output stride $16$ on the PASCAL VOC dataset. The iteration number $T$ for training is set as $3$. \textit{Best viewed on screen}.}
		\label{init-norm}
		\vspace{-5mm}
	\end{figure}
	
	\section{Experiments}
	
	To evaluate the proposed EMAU, we conduct extensive experiments on the PASCAL VOC dataset~\cite{voc}, the PASCAL Context dataset~\cite{pcontext}, and the COCO Stuff dataset~\cite{coco}. 
	In this section, we first introduce implementation details. Then we perform ablation study to verify the superiority of proposed method on the PASCAL VOC dataset. Finally, we report our results on the PASCAL Context dataset and the COCO Stuff dataset.

	\subsection{Implementation Details}
	\vspace{1mm}
	We use ResNet~\cite{resnet}~(pretrained on ImageNet~\cite{imagenet}) as our backbone. Following prior works~\cite{pspnet,deeplabv3,deeplabv3+}, we employ a poly learning rate policy where the initial learning rate is multiplied by $ \left( 1 -  \mathrm{iter} / \mathrm{total\_iter} \right)^{0.9} $ after each iteration. The initial learning rate is set to be $0.009$ for all datasets. Momentum and weight decay coefficients are set to $0.9$ and $0.0001$, respectively. For data augmentation, we apply the common scale~($0.5$ to $2.0$), cropping and  flipping of the image to augment the training data. Input size for all datasets is set to $513 \times 513$. The synchronized batch normalization is adopted in all experiments, together with the multi-grid~\cite{deeplabv3}. For evaluation, we adopt the commonly used Mean IoU metric. 
	
	The output stride of the backbone is set to $16$ for training on PASCAL VOC and PASCAL Context, and $8$ for training on COCO Stuff and evaluating on all datasets. To speed up the training procedure, we carry out all ablation studies on ResNet-50~\cite{resnet}, with batch size $12$. For all models to be compared with state-of-the-art, we train them on ResNet-101, with batch size $16$. We train 30K iterations on PASCAL VOC and COCO Stuff, and 15K on PASCAL Context. We use a $3 \times 3$ convolution to reduce the channel number from $2,048$ to $512$, and then stack EMAU upon it. We call the whole network as \textbf{EMANet}. We set the basis number $K = 64$, $\lambda = 1$ and the number of iterations $T = 3$ for training as default. 
	
	\begin{figure}
		\centering
		\includegraphics[width=0.48\textwidth]{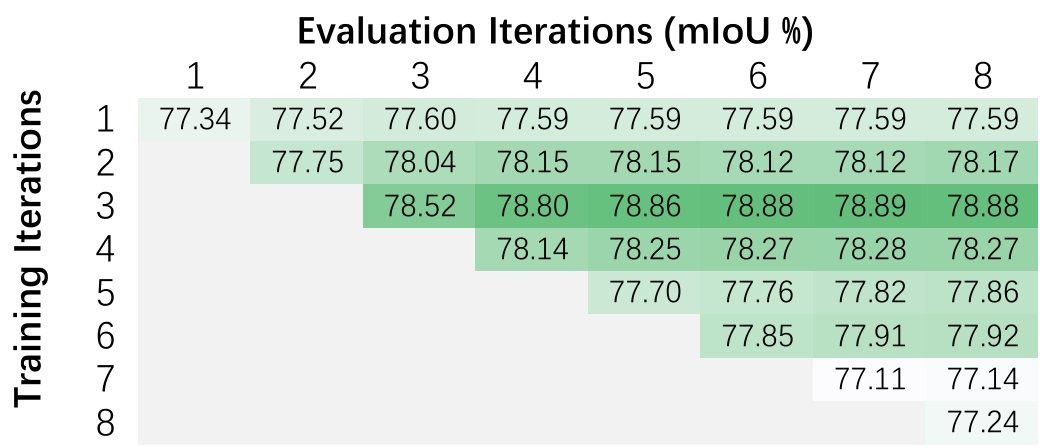}
		\vspace{-2mm}
		\caption{Ablation study on the iteration number $T$. Experiments are conducted upon ResNet-50 with training output stride $16$ and batch size $12$ on the PASCAL VOC dataset.}
		\label{fig-T}
		\vspace{-2mm}
	\end{figure}
	
	\subsection{Results on the PASCAL VOC Dataset} 
	\vspace{2mm}
	\subsubsection{Bases Maintenance and Normalization}
	\vspace{1mm}
	In this part, we first compare different strategies of maintaining $\bm{\mu}^{(0)}$. We set $T = 3$ in training, and $1 \leq T \leq 8$ in evaluation. As shown in the left part of Fig.~\ref{init-norm}, performance of all strategies increases with more iterations of $A_{\mathrm{E}}$ and $A_{\mathrm{M}}$. When $T \geq 4$, the gain from more iterations becomes marginal. \textbf{Moving average} performs the best among them. It achieves the highest performances in all iterations and surpasses others by at least $0.9$ in mIoU. Surprisingly, updating by the back propagation shows no merit compared with no updating and even performs worse when $T \geq 3$. 
	
	We then compare the performances with no normalization, LN and L2Norm as described above. From the right part of Fig.~\ref{init-norm}, it is clear to see that LN is even worse than no normalization. Since it can partially relieve the gradient chores of RNN-like structure. The performance of LN and no normalization has little correlation with the number of iteration $T$. By contrast, L2Norm's performance increases as the iterations become larger and it outperforms LN and no normalization when $T \geq 3$.

	\begin{table}[t!]
		\caption{Detailed comparisons on PASCAL VOC with DeeplabV3/V3+ and PSANet in mIoU (\%). All results are achieved with the backbone ResNet-101 and output stride 8. The FLOPs and memory are computed with the input size $513 \times 513$. \textbf{SS}: Single scale input during test. \textbf{MS}: Multi-scale input. \textbf{Flip}: Adding left-right flipped input. EMANet~(256) and EMANet~(512) represent EMANet with the number of input channels as 256 and 512, respectively.}
		\vspace{-2mm}
		\label{tab-voc-detail}
		\centering
		\small
		\setlength\tabcolsep{1.1mm}
		\begin{tabular}{l|cc|ccc}
			\hline
			Method     					& SS    & MS+Flip & FLOPs  & Memory & Params 	\\ \hline \hline
			ResNet-101   				& -	    & -		  & 190.6G & 2.603G	& 42.6M		\\ \hline
			DeeplabV3~\cite{deeplabv3}	& 78.51 & 79.77   & +63.4G & +66.0M	& +15.5M		\\
			DeeplabV3+~\cite{deeplabv3+}& 79.35 & 80.57   & +84.1G & +99.3M & +16.3M		\\
			PSANet~\cite{psanet}		& 78.51 & 79.77	  & +56.3G & +59.4M & +18.5M		\\
			\textbf{EMANet}~(256)     	& \underline{79.73} & \underline{80.94}   & \textbf{+21.1G} & \textbf{+12.3M}  & \textbf{+4.87M}		\\
			\textbf{EMANet}~(512)     	& \textbf{80.05} & \textbf{81.32}   & \underline{+43.1G} & \underline{+22.1M}  & \underline{+10.0M}	\\ \hline
		\end{tabular}
		\vspace{-2mm}
	\end{table}
	
	\begin{table}[t!]
	\caption{Comparisons on the PASCAL VOC test set.}
	\vspace{-2mm}
	\label{tab-voc-sota}
	\centering
	\setlength\tabcolsep{3.5mm}
	\begin{tabular}{l|c|c}
		\hline
		Method          			& Backbone      & mIoU~(\%) \\ \hline\hline
		Wide ResNet~\cite{wider} 	& WideResNet-38 & 84.9 \\
		PSPNet~\cite{pspnet} 		& ResNet-101    & 85.4 \\
		DeeplabV3~\cite{deeplabv3}	& ResNet-101 	& 85.7 \\
		PSANet~\cite{psanet}		& ResNet-101	& 85.7 \\
		EncNet~\cite{encnet} 		& ResNet-101    & 85.9 \\
		DFN~\cite{dfn}				& ResNet-101    & 86.2 \\
		Exfuse~\cite{exfuse}		& ResNet-101    & 86.2 \\
		IDW-CNN~\cite{idn}         	& ResNet-101    & 86.3 \\
		SDN~\cite{sdn} 	& DenseNet-161  & 86.6 \\
		DIS~\cite{dis}             	& ResNet-101    & 86.8 \\
		\textbf{EMANet} & ResNet-101& \textbf{87.7} \\ \hline
		GCN~\cite{gcn}     			& ResNet-152    & 83.6 \\
		RefineNet~\cite{refinenet}	& ResNet-152    & 84.2 \\
		DeeplabV3+~\cite{deeplabv3+}& Xception-71   & 87.8 \\
		Exfuse~\cite{exfuse} 		& ResNeXt-131   & 87.9 \\
		MSCI~\cite{msci}            & ResNet-152    & 88.0 \\ 
		\textbf{EMANet} & ResNet-152& \textbf{88.2} \\ \hline
	\end{tabular}
	\vspace{-2mm}
    \end{table}
	
	\subsubsection{Ablation Study for Iteration Number}\label{sec-iter}
	
	From Fig.~\ref{init-norm}, it is obvious that the performance of EMAU gain from more iterations during evaluation, and the gain becomes marginal when $T > 4$. In this subsection, we also study the influence of $T$ in training. We plot the performance matrix upon $T_{\mathrm{train}}$ and $T_{\mathrm{eval}}$ as Fig.~\ref{fig-T}.


	From Fig.~\ref{fig-T}, it is clear that mIoU increases monotonically with more iterations in evaluation, no matter what $T_{\mathrm{train}}$ is. They finally converge to a fixed value. However, this rule does not work in training. The mIoUs peak when $T_{\mathrm{train}} = 3$ and decrease with more iterations. This phenomenon may be caused by the RNN-like behavior of EMAU. Though Moving Average and L2Norm can relieve to a certain degree, the problem persists.
	
	We also carry out experiments on $A^2$ block~\cite{a2net}, which can be regarded as a special form of EMAU as mentioned in Sec.~\ref{sec-a2net}. Similarly, the non-local module can also be viewed as a special form of EMAU without $A_{\mathrm{M}}$ step, which includes more bases and $T_{\mathrm{train}} = 1$. With the same backbone and training scheduler, $A^2$ block achieves 77.41\% and the non-local module  achieves 77.78\% in mIoU, respectively. As a comparison, EMANet achieves 77.34\% when $T_{\mathrm{train}}=1$ and $T_{\mathrm{eval}}=1$. These three results have small differences, which is coincident with our analysis.

\begin{table}[t]
	\caption{Comparisons with state-of-the-art on the PASCAL Context test set. `+' means pretrained on COCO Stuff.}
	\label{tab-pcontext}
	\centering
	\setlength\tabcolsep{5.8mm}
	\vspace{-2mm}
	\begin{tabular}{l|c|c}
		\hline
		Method      			& Backbone 		  & mIoU~(\%) \\ \hline\hline
		PSPNet~\cite{pspnet} 	& ResNet-101      & 47.8 \\
		DANet~\cite{danet}   	& ResNet-50       & 50.1 \\
		MSCI~\cite{msci}     	& ResNet-152      & 50.3 \\
		\textbf{EMANet}      	& ResNet-50       & \underline{50.5} \\
		SGR~\cite{sgr}	       	& ResNet-101      & 50.8 \\
		CCL~\cite{ccl}       	& ResNet-101      & 51.6 \\
		EncNet~\cite{encnet} 	& ResNet-101      & 51.7 \\
		SGR+~\cite{sgr}	       	& ResNet-101      & 52.5 \\
		DANet~\cite{danet}      & ResNet-101      & 52.6 \\
		\textbf{EMANet}      	& ResNet-101      & \textbf{53.1} \\ \hline
	\end{tabular}
	\vspace{-2mm}
\end{table}

\begin{table}[t]
	\caption{Comparisons on the COCO Stuff test set.}
	\label{tab-coco}
	\centering
	\setlength\tabcolsep{5.2mm}
	\vspace{-2mm}
	\begin{tabular}{l|c|c}
		\hline
		Method         				& Backbone 		& mIoU~(\%) \\ \hline\hline
		RefineNet~\cite{refinenet}	& ResNet-101    & 33.6 \\
		CCL~\cite{ccl}				& ResNet-101    & 35.7 \\
		DANet~\cite{danet}  		& ResNet-50     & 37.2 \\
		DSSPN~\cite{dsspn}          & ResNet-101    & 37.3 \\
		\textbf{EMANet}				& ResNet-50     & \underline{37.6} \\
		SGR~\cite{sgr}            	& ResNet-101    & 39.1 \\
		DANet~\cite{danet}          & ResNet-101    & 39.7 \\
		\textbf{EMANet}         	& ResNet-101    & \textbf{39.9} \\ \hline
	\end{tabular}
	\vspace{-2mm}
\end{table}
	
\subsubsection{Comparisons with State-of-the-arts}

\begin{figure*}[t!]
	\centering
	\begin{subfigure}{0.16\textwidth}
		\includegraphics[width=\textwidth]{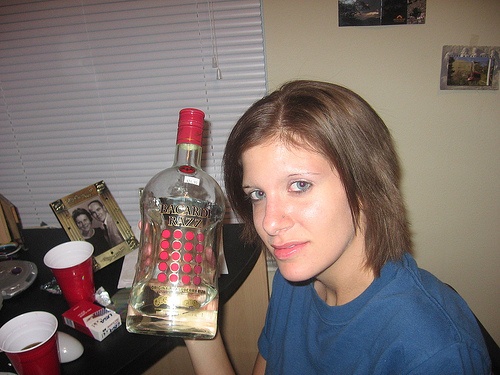}
	\end{subfigure}
	\begin{subfigure}{0.16\textwidth}
		\includegraphics[width=\textwidth]{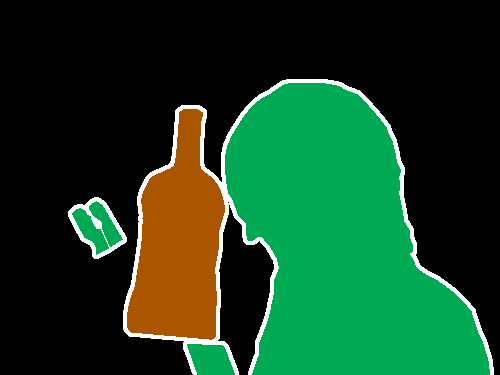}
	\end{subfigure}
	\begin{subfigure}{0.16\textwidth}
		\includegraphics[width=\textwidth]{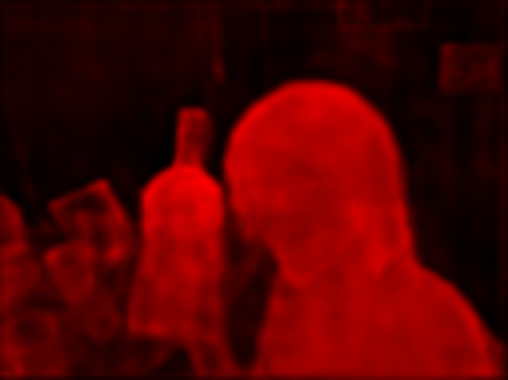}
	\end{subfigure}
	\begin{subfigure}{0.16\textwidth}
		\includegraphics[width=\textwidth]{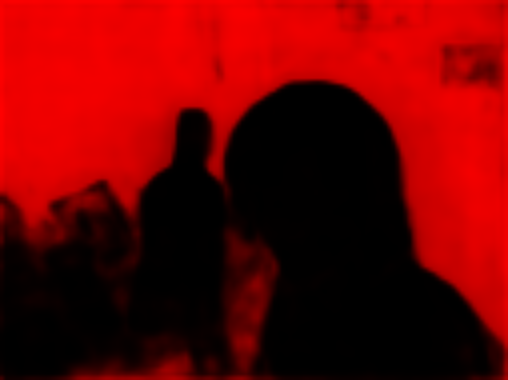}
	\end{subfigure}
	\begin{subfigure}{0.16\textwidth}
		\includegraphics[width=\textwidth]{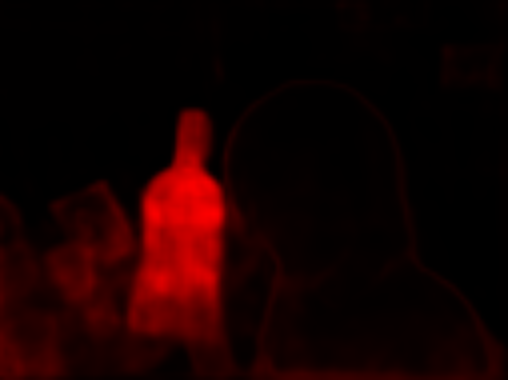}
	\end{subfigure}
	\begin{subfigure}{0.16\textwidth}
		\includegraphics[width=\textwidth]{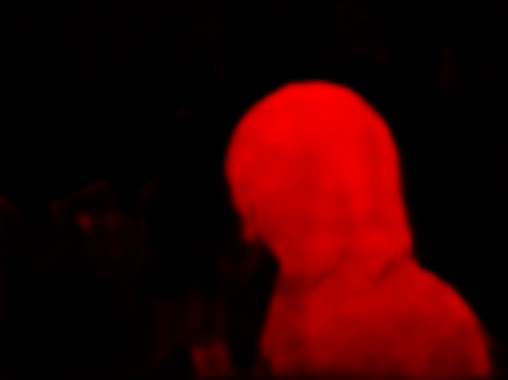}
	\end{subfigure}
	
	\begin{subfigure}{0.16\textwidth}
		\includegraphics[width=\textwidth]{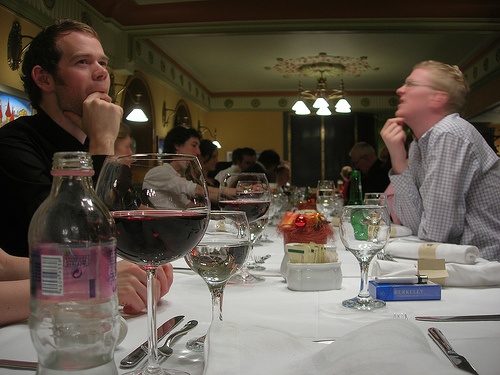}
	\end{subfigure}
	\begin{subfigure}{0.16\textwidth}
		\includegraphics[width=\textwidth]{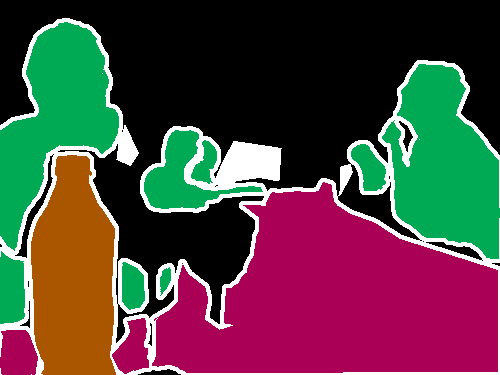}
	\end{subfigure}
	\begin{subfigure}{0.16\textwidth}
		\includegraphics[width=\textwidth]{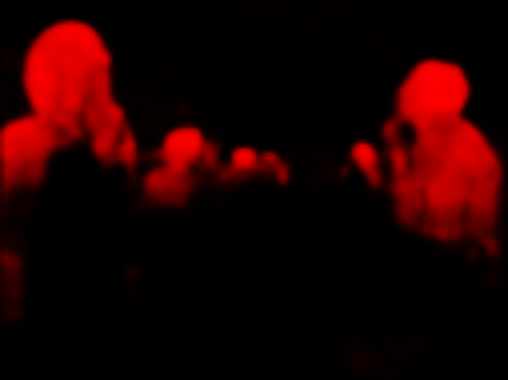}
	\end{subfigure}
	\begin{subfigure}{0.16\textwidth}
		\includegraphics[width=\textwidth]{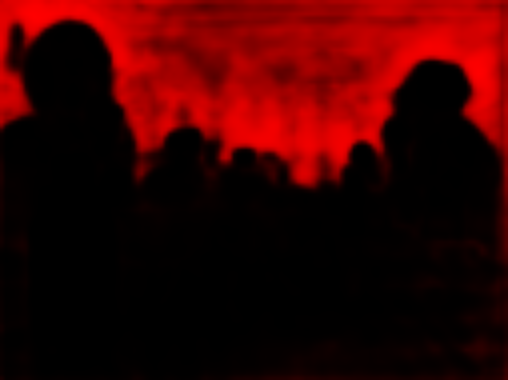}
	\end{subfigure}
	\begin{subfigure}{0.16\textwidth}
		\includegraphics[width=\textwidth]{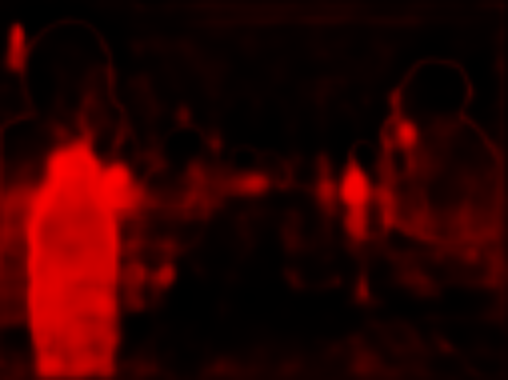}
	\end{subfigure}
	\begin{subfigure}{0.16\textwidth}
		\includegraphics[width=\textwidth]{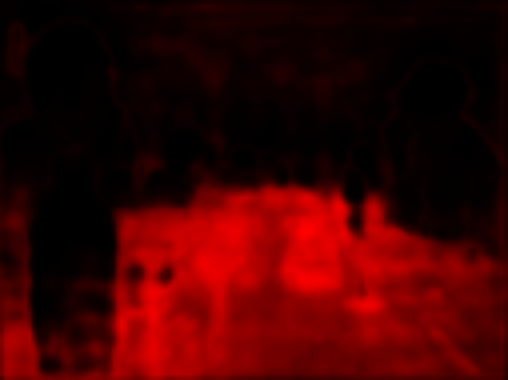}
	\end{subfigure}

	\begin{subfigure}{0.16\textwidth}
		\includegraphics[width=\textwidth]{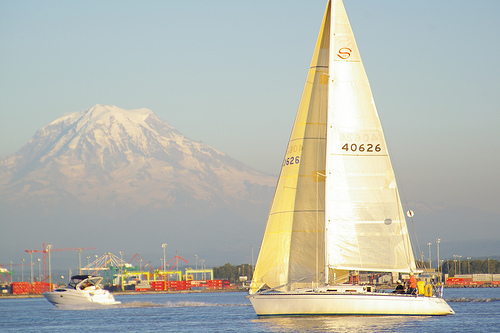}
	\end{subfigure}
	\begin{subfigure}{0.16\textwidth}
		\includegraphics[width=\textwidth]{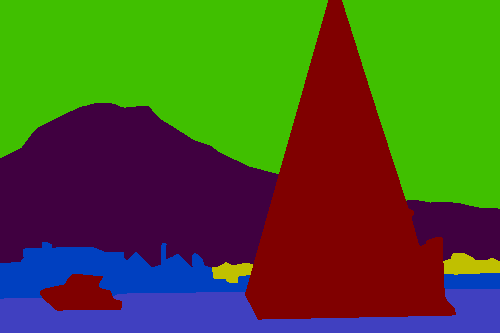}
	\end{subfigure}
	\begin{subfigure}{0.16\textwidth}
		\includegraphics[width=\textwidth]{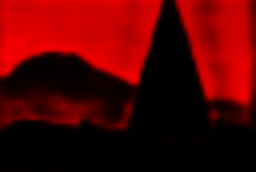}
	\end{subfigure}
	\begin{subfigure}{0.16\textwidth}
		\includegraphics[width=\textwidth]{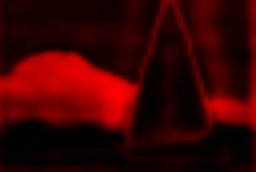}
	\end{subfigure}
	\begin{subfigure}{0.16\textwidth}
		\includegraphics[width=\textwidth]{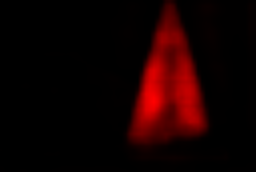}
	\end{subfigure}
	\begin{subfigure}{0.16\textwidth}
		\includegraphics[width=\textwidth]{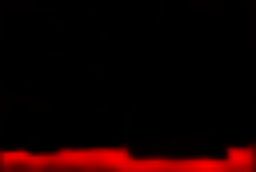}
	\end{subfigure}
	
	\begin{subfigure}{0.16\textwidth}
		\includegraphics[width=\textwidth]{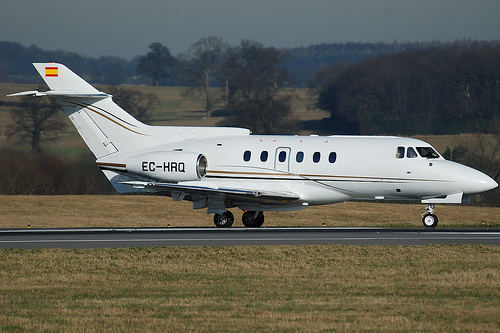}
		\caption{Image}
	\end{subfigure}
	\begin{subfigure}{0.16\textwidth}
		\includegraphics[width=\textwidth]{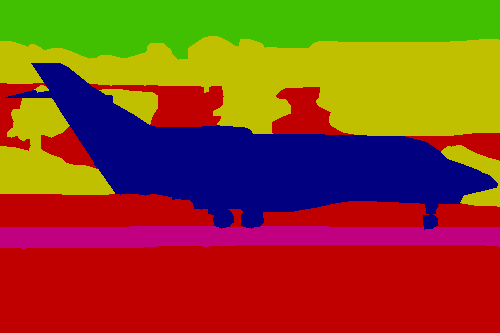}
		\caption{Label}
	\end{subfigure}
	\begin{subfigure}{0.16\textwidth}
		\includegraphics[width=\textwidth]{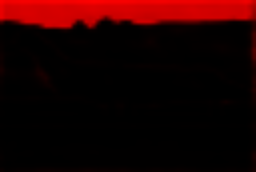}
		\caption{\textbf{$z_{\cdot i}$}}
	\end{subfigure}
	\begin{subfigure}{0.16\textwidth}
		\includegraphics[width=\textwidth]{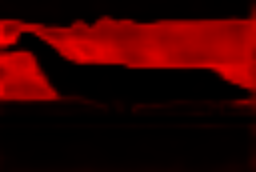}
		\caption{\textbf{$z_{\cdot j}$}}
	\end{subfigure}
	\begin{subfigure}{0.16\textwidth}
		\includegraphics[width=\textwidth]{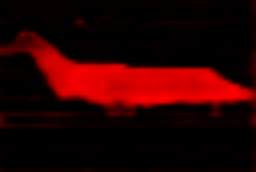}
		\caption{\textbf{$z_{\cdot k}$}}
	\end{subfigure}
	\begin{subfigure}{0.16\textwidth}
		\includegraphics[width=\textwidth]{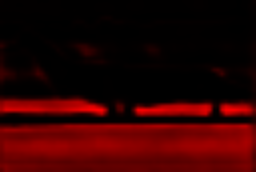}
		\caption{\textbf{$z_{\cdot l}$}}
	\end{subfigure}
	
	\vspace{-1mm}
	\caption{Visualization of responsibilities $\mathbf{Z}$ at the last iteration. The first two rows illustrate two examples from the PASCAL VOC validation set. The last two rows illustrate two examples from the PASCAL Context validation set. $z_{\cdot i}$ represents the responsibilities of the $i$-th basis to all pixels in the last iteration, $i, j, k$ and $l$ are four \textit{randomly} selected indexes, where $1 \leq i, j, k, l \leq K$. \textit{Best viewed on screen}.}
	\label{fig-Z}
	\vspace{-3mm}
\end{figure*}
	

We first thoroughly compare EMANet with three baselines, namely DeeplabV3, DeeplabV3+ and PSANet on the validation set. We report mIoU, FLOPs, memory cost and parameter numbers in Tab.~\ref{tab-voc-detail}. We can see that EMANet outperforms these three baselines by a large margin. Moreover, EMANet is much lighter in computation and memory.

We further compare our method with existing methods on the PASCAL VOC test set. Following previous methods~\cite{deeplabv3,deeplabv3+}, we train EMANet successively over COCO, the VOC \textit{trainaug} and the VOC \textit{trainval} set. We set the base learning rate as 0.009, 0.001 and 0.0001, respectively. We train 150K iterations on COCO, and 30K for the last two rounds. When inferring over the \textit{test} set, we make use of multi-scale testing and left-right flipping.

As shown in Tab.~\ref{tab-voc-sota}, our EMANet sets the new record on PASCAL VOC, and improves DeeplabV3~\cite{deeplabv3} with the same backbone by \textbf{2.0\%} in mIoU. Our EMANet achieves the best performance among networks with backbone ResNet-101, and outperforms the previous best one by \textbf{0.9\%}, which is significant due to the fact that this benchmark is very competitive. Moreover, it achieves the performance that is comparable with methods based on some larger backbones.

\subsection{Results on the PASCAL Context Dataset}

To verify the generalization of our proposed EMANet, we conduct experiments on the PASCAL Context dataset. Quantitative results of PASCAL Context are shown in Tab.~\ref{tab-pcontext}. To the best of our knowledge, EMANet based on ResNet-101 achieves the highest performance on the PASCAL Context dataset. Even pretrained on additional data~(COCO Stuff), SGR+ is still inferior to EMANet.

\subsection{Results on the COCO Stuff Dataset}

To further evaluate the effectiveness of our method, we also carry out experiments on the  COCO Stuff dataset. Comparisons with previous state-of-the-art methods are shown in Tab.~\ref{tab-coco}. Remarkably,  EMANet achieves 39.9\% in mIoU and outperforms previous methods by a large margin.

\subsection{Visualization of Bases Responsibilities}\label{sec-vis}
	
	To get a deeper understanding of our proposed EMAU, we visualize the iterated responsibility map $\mathbf{Z}$ in Fig.~\ref{fig-Z}. For each image, we randomly select four bases~($i, j, k$ and $l$) and show their corresponding responsibilities of all pixels in the last iteration. Obviously, each basis corresponds to an abstract concept of the image. 
	With the progress of iterations $A_{\mathrm E}$ and $A_{\mathrm M}$, the abstract concept becomes more compact and clear. 
	As we can see, these bases converge to some specific semantics and do not just focus on foreground and background. 
	Concretely, the bases of the first two rows focus on specific semantics such as human, wine glass, cutlery and profile. The bases of the last two rows focus on sailboat, mountain, airplane and lane.

\section{Conclusion}

In this paper, we propose a new type of attention mechanism, namely the expectation-maximization attention~(EMA), which computes a more compact basis set by iteratively executing as the EM algorithm. The reconstructed output of EMA is low-rank and robust to the variance of input. We well formulate the proposed method as a light-weighted module that can be easily inserted to existing CNNs with little overhead. Extensive experiments on a number of benchmark datasets demonstrate the effectiveness and efficiency of the proposed EMAU.

\section*{Acknowledgment}

Zhouchen Lin is supported by National Basic Research Program of China (973 Program) (Grant no. 2015CB352502), National Natural Science Foundation (NSF) of China (Grant nos. 61625301 and 61731018), Qualcomm and Microsoft Research Asia. Hong Liu is supported by National Natural Science Foundation of China (Grant nos. U1613209 and 61673030) and funds from Shenzhen Key Laboratory for Intelligent Multimedia and Virtual Reality (ZDSYS201703031405467).

{\small
	\bibliographystyle{ieee_fullname}
	\bibliography{egbib}

\begin{thebibliography}{10}\itemsep=-1pt

\bibitem{ln}
Jimmy~Lei Ba, Jamie~Ryan Kiros, and Geoffrey~E Hinton.
\newblock Layer normalization.
\newblock {\em arXiv preprint arXiv:1607.06450}, 2016.

\bibitem{nmt}
Dzmitry Bahdanau, Kyunghyun Cho, and Yoshua Bengio.
\newblock Neural machine translation by jointly learning to align and
  translate.
\newblock {\em arXiv preprint arXiv:1409.0473}, 2014.

\bibitem{coco}
Holger Caesar, Jasper Uijlings, and Vittorio Ferrari.
\newblock Coco-stuff: Thing and stuff classes in context.
\newblock In {\em CVPR}, pages 1209--1218, 2018.

\bibitem{deeplabv3}
Liang-Chieh Chen, George Papandreou, Florian Schroff, and Hartwig Adam.
\newblock Rethinking atrous convolution for semantic image segmentation.
\newblock {\em arXiv preprint arXiv:1706.05587}, 2017.

\bibitem{deeplabv3+}
Liang-Chieh Chen, Yukun Zhu, George Papandreou, Florian Schroff, and Hartwig
  Adam.
\newblock Encoder-decoder with atrous separable convolution for semantic image
  segmentation.
\newblock In {\em ECCV}, 2018.

\bibitem{a2net}
Yunpeng Chen, Yannis Kalantidis, Jianshu Li, Shuicheng Yan, and Jiashi Feng.
\newblock A2-nets: Double attention networks.
\newblock In {\em NeurIPS}, pages 350--359, 2018.

\bibitem{em}
Arthur~P Dempster, Nan~M Laird, and Donald~B Rubin.
\newblock Maximum likelihood from incomplete data via the em algorithm.
\newblock {\em Journal of the Royal Statistical Society: Series B
  (Methodological)}, 39(1):1--22, 1977.

\bibitem{ccl}
Henghui Ding, Xudong Jiang, Bing Shuai, Ai Qun~Liu, and Gang Wang.
\newblock Context contrasted feature and gated multi-scale aggregation for
  scene segmentation.
\newblock In {\em CVPR}, pages 2393--2402, 2018.

\bibitem{voc}
Mark Everingham, Luc Van~Gool, Christopher~KI Williams, John Winn, and Andrew
  Zisserman.
\newblock The pascal visual object classes (voc) challenge.
\newblock {\em International journal of computer vision}, 88(2):303--338, 2010.

\bibitem{kmeans}
Edward~W Forgy.
\newblock Cluster analysis of multivariate data: efficiency versus
  interpretability of classifications.
\newblock {\em biometrics}, 21:768--769, 1965.

\bibitem{danet}
Jun Fu, Jing Liu, Haijie Tian, Yong Li, Yongjun Bao, Zhiwei Fang, and Hanqing
  Lu.
\newblock Dual attention network for scene segmentation.
\newblock In {\em CVPR}, pages 3146--3154, 2019.

\bibitem{sdn}
Jun Fu, Jing Liu, Yuhang Wang, Jin Zhou, Changyong Wang, and Hanqing Lu.
\newblock Stacked deconvolutional network for semantic segmentation.
\newblock {\em IEEE Transactions on Image Processing}, 2019.

\bibitem{kaiming}
Kaiming He, Xiangyu Zhang, Shaoqing Ren, and Jian Sun.
\newblock Delving deep into rectifiers: Surpassing human-level performance on
  imagenet classification.
\newblock In {\em ICCV}, pages 1026--1034, 2015.

\bibitem{resnet}
Kaiming He, Xiangyu Zhang, Shaoqing Ren, and Jian Sun.
\newblock Deep residual learning for image recognition.
\newblock In {\em CVPR}, pages 770--778, 2016.

\bibitem{densenet}
Gao Huang, Zhuang Liu, Laurens Van Der~Maaten, and Kilian~Q Weinberger.
\newblock Densely connected convolutional networks.
\newblock In {\em CVPR}, pages 4700--4708, 2017.

\bibitem{bn}
Sergey Ioffe and Christian Szegedy.
\newblock Batch normalization: Accelerating deep network training by reducing
  internal covariate shift.
\newblock {\em arXiv preprint arXiv:1502.03167}, 2015.

\bibitem{rescan}
Xia Li, Jianlong Wu, Zhouchen Lin, Hong Liu, and Hongbin Zha.
\newblock Recurrent squeeze-and-excitation context aggregation net for single
  image deraining.
\newblock In {\em ECCV}, pages 254--269, 2018.

\bibitem{sgr}
Xiaodan Liang, Zhiting Hu, Hao Zhang, Liang Lin, and Eric~P Xing.
\newblock Symbolic graph reasoning meets convolutions.
\newblock In {\em NeurIPS}, pages 1858--1868, 2018.

\bibitem{dsspn}
Xiaodan Liang, Hongfei Zhou, and Eric Xing.
\newblock Dynamic-structured semantic propagation network.
\newblock In {\em CVPR}, pages 752--761, 2018.

\bibitem{msci}
Di Lin, Yuanfeng Ji, Dani Lischinski, Daniel Cohen-Or, and Hui Huang.
\newblock Multi-scale context intertwining for semantic segmentation.
\newblock In {\em ECCV}, pages 603--619, 2018.

\bibitem{refinenet}
Guosheng Lin, Anton Milan, Chunhua Shen, and Ian Reid.
\newblock Refinenet: Multi-path refinement networks for high-resolution
  semantic segmentation.
\newblock In {\em CVPR}, pages 1925--1934, 2017.

\bibitem{fcn}
Jonathan Long, Evan Shelhamer, and Trevor Darrell.
\newblock Fully convolutional networks for semantic segmentation.
\newblock In {\em CVPR}, pages 3431--3440, 2015.

\bibitem{dis}
Ping Luo, Guangrun Wang, Liang Lin, and Xiaogang Wang.
\newblock Deep dual learning for semantic image segmentation.
\newblock In {\em ICCV}, pages 2718--2726, 2017.

\bibitem{pcontext}
Roozbeh Mottaghi, Xianjie Chen, Xiaobai Liu, Nam-Gyu Cho, Seong-Whan Lee, Sanja
  Fidler, Raquel Urtasun, and Alan Yuille.
\newblock The role of context for object detection and semantic segmentation in
  the wild.
\newblock In {\em CVPR}, pages 891--898, 2014.

\bibitem{gcn}
Chao Peng, Xiangyu Zhang, Gang Yu, Guiming Luo, and Jian Sun.
\newblock Large kernel matters--improve semantic segmentation by global
  convolutional network.
\newblock In {\em CVPR}, pages 4353--4361, 2017.

\bibitem{gmm}
Sylvia Richardson and Peter~J Green.
\newblock On bayesian analysis of mixtures with an unknown number of components
  (with discussion).
\newblock {\em Journal of the Royal Statistical Society: series B (statistical
  methodology)}, 59(4):731--792, 1997.

\bibitem{unet}
Olaf Ronneberger, Philipp Fischer, and Thomas Brox.
\newblock U-net: Convolutional networks for biomedical image segmentation.
\newblock In {\em MICCAI}, pages 234--241. Springer, 2015.

\bibitem{imagenet}
Olga Russakovsky, Jia Deng, Hao Su, Jonathan Krause, Sanjeev Satheesh, et~al.
\newblock Imagenet large scale visual recognition challenge.
\newblock {\em IJCV}, 115(3):211--252, 2015.

\bibitem{attention}
Ashish Vaswani, Noam Shazeer, Niki Parmar, Jakob Uszkoreit, Llion Jones,
  Aidan~N Gomez, {\L}ukasz Kaiser, and Illia Polosukhin.
\newblock Attention is all you need.
\newblock In {\em NeurIPS}, pages 5998--6008, 2017.

\bibitem{idn}
Guangrun Wang, Ping Luo, Liang Lin, and Xiaogang Wang.
\newblock Learning object interactions and descriptions for semantic image
  segmentation.
\newblock In {\em CVPR}, pages 5859--5867, 2017.

\bibitem{nonlocal}
Xiaolong Wang, Ross Girshick, Abhinav Gupta, and Kaiming He.
\newblock Non-local neural networks.
\newblock In {\em CVPR}, pages 7794--7803, 2018.

\bibitem{wider}
Zifeng Wu, Chunhua Shen, and Anton Van Den~Hengel.
\newblock Wider or deeper: Revisiting the resnet model for visual recognition.
\newblock {\em Pattern Recognition}, 90:119--133, 2019.

\bibitem{cliquenet}
Yibo Yang, Zhisheng Zhong, Tiancheng Shen, and Zhouchen Lin.
\newblock Convolutional neural networks with alternately updated clique.
\newblock In {\em CVPR}, pages 2413--2422, 2018.

\bibitem{dfn}
Changqian Yu, Jingbo Wang, Chao Peng, Changxin Gao, Gang Yu, and Nong Sang.
\newblock Learning a discriminative feature network for semantic segmentation.
\newblock In {\em CVPR}, pages 1857--1866, 2018.

\bibitem{encnet}
Hang Zhang, Kristin Dana, Jianping Shi, Zhongyue Zhang, Xiaogang Wang, Ambrish
  Tyagi, and Amit Agrawal.
\newblock Context encoding for semantic segmentation.
\newblock In {\em CVPR}, pages 7151--7160, 2018.

\bibitem{exfuse}
Zhenli Zhang, Xiangyu Zhang, Chao Peng, Xiangyang Xue, and Jian Sun.
\newblock Exfuse: Enhancing feature fusion for semantic segmentation.
\newblock In {\em ECCV}, pages 269--284, 2018.

\bibitem{pspnet}
Hengshuang Zhao, Jianping Shi, Xiaojuan Qi, Xiaogang Wang, and Jiaya Jia.
\newblock Pyramid scene parsing network.
\newblock In {\em CVPR}, pages 2881--2890, 2017.

\bibitem{psanet}
Hengshuang Zhao, Yi Zhang, Shu Liu, Jianping Shi, Chen Change~Loy, Dahua Lin,
  and Jiaya Jia.
\newblock Psanet: Point-wise spatial attention network for scene parsing.
\newblock In {\em ECCV}, pages 267--283, 2018.

\end{thebibliography}
}

\end{document}